\documentclass[sigconf]{acmart}
\renewcommand\footnotetextcopyrightpermission[1]{} % removes footnote with conference information in first column
\AtBeginDocument{%
  \providecommand\BibTeX{{%
    \normalfont B\kern-0.5em{\scshape i\kern-0.25em b}\kern-0.8em\TeX}}}

\acmConference[]{}
\usepackage{soul}
\usepackage{url}
\usepackage[utf8]{inputenc}
\usepackage{caption}
\usepackage{graphicx}
\usepackage{amsmath}
\usepackage{booktabs}
\usepackage{multirow}
\usepackage{tabularx}
\usepackage{enumitem}
\usepackage{float}
\usepackage{subfloat}
\usepackage{subfig}
\graphicspath{{figures/}}
\usepackage[ruled,linesnumbered]{algorithm2e}

\newcolumntype{L}[1]{>{\raggedright\let\newline\\\arraybackslash\hspace{0pt}}m{#1}}
\newcolumntype{C}[1]{>{\centering\let\newline\\\arraybackslash\hspace{0pt}}m{#1}}
\newcolumntype{R}[1]{>{\raggedleft\let\newline\\\arraybackslash\hspace{0pt}}m{#1}}

\newcolumntype{R}{>{\raggedleft\arraybackslash}X}
\newcolumntype{C}{>{\centering\arraybackslash}X}
\usepackage{array}
\begin{document}

\title{TCDesc: Learning Topology Consistent Descriptors}


% \author{Anonymous}

\author{Honghu Pan}
\email{19b951002@stu.hit.edu.cn}
\affiliation{%
	\institution{Harbin Institute of Technology, Shenzhen}
	\city{Shenzhen}
	\country{China}
}

\author{Fanyang Meng}
\email{mengfy@pcl.ac.cn}
\affiliation{%
	\institution{Peng Cheng Laboratory}
	\city{Shenzhen}
	\country{China}}

\author{Zhenyu He}
\email{zhenyuhe@hit.edu.cn}
\affiliation{%
	\institution{Harbin Institute of Technology, Shenzhen}
	\city{Shenzhen}
	\country{China}
}

\author{Yongsheng Liang}
\email{liangys@hit.edu.cn}
\affiliation{%
	\institution{Harbin Institute of Technology, Shenzhen}
	\city{Shenzhen}
	\country{China}
}

\author{Wei Liu}
\email{liuwei@sziit.edu.cn}
\affiliation{%
	\institution{Peng Cheng Laboratory}
	\city{Shenzhen}
	\country{China}}

% \affiliation{%



% \renewcommand{\shortauthors}{Anonymous}

\begin{abstract}
Triplet loss is widely used for learning local descriptors from image patch. However, triplet loss only minimizes the Euclidean distance between matching descriptors and maximizes that between the non-matching descriptors, which neglects the topology similarity between two descriptor sets. In this paper, we propose topology measure besides Euclidean distance to learn topology consistent descriptors by considering $k$NN descriptors of positive sample. First we establish a novel topology vector for each descriptor followed by Locally Linear Embedding (LLE) to indicate the topological relation among the descriptor and its $k$NN descriptors. Then we define topology distance between descriptors as the difference of their topology vectors. 
Last we employ the dynamic weighting strategy to fuse Euclidean distance and topology distance of matching descriptors and take the fusion result as the positive sample distance in the triplet loss. Experimental results on several benchmarks show that our method performs better than state-of-the-arts results and effectively improves the performance of triplet loss.
\end{abstract}


\maketitle

\section{Introduction}

Image matching is a fundamental computer vision problem and the crucial step in augmented reality(AR)~\cite{ARsurvey,markerlessAR} and simultaneous localization and mapping(SLAM)~\cite{ORB-SLAM,ORB-SLAM2}, which usually consists of two steps: detecting the feature points and matching feature descriptors. The robust and discriminative descriptors are essential for accurate image matching.
% handcrafted descriptors
Early works mainly focus on the handcrafted descriptors.
SIFT~\cite{SIFT} maybe is the most successful handcrafted descriptor which has been proven effective in various areas~\cite{csurka2004visual,zheng2017sift,wide_baseline}.
%and widely used in image matching, image retrieval~\cite{csurka2004visual,zheng2017sift} and wide baseline stereo~\cite{wide_baseline}. 
Meanwhile, the binary descriptors~\cite{Brief} are proposed to reduce storage and accelerate matching.
However, handcrafted descriptors are not robust enough due to the lack of high-level semantic information. 

Recently with the successful application of CNN in multiple fields~\cite{lecun2015deep,girshick2015fast,bertinetto2016fully}, researchers~\cite{MatchNet,DeepDesc,DeepCompare,L2Net,HardNet} try to learn descriptors directly from image patch by using CNN. 
Recent works~\cite{HardNet,ExpLoss,DSM} mainly focus on learning descriptors using triplet loss~\cite{FaceNet} to encourage Euclidean distance of negtive samples is a margin larger than that of positive samples, where negtive samples and positive samples denote the non-matching descriptors and matching descriptors respectively.
Specifically, CNN takes two image patch sets with one-to-one matching relationship as input and outputs corresponding two descriptors sets, where the Euclidean distance of matching descriptors is minimized and that of non-matching descriptors is maximized.

\begin{figure}[t]
	% \vspace{-5pt}
	\centering
	\renewcommand{\thesubfigure}{a}
	\subfloat[Former triplet loss \label{distribution-comparison-a}]{%
		\includegraphics[width=0.23 \textwidth]{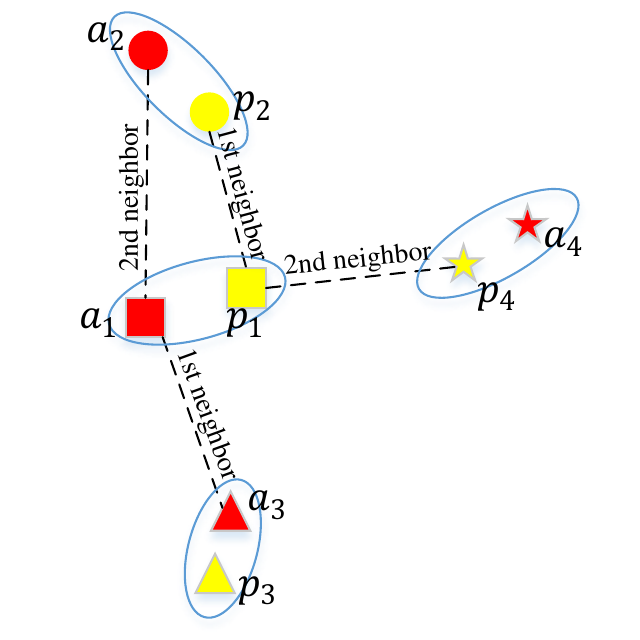}
	}
	\hfill
	\renewcommand{\thesubfigure}{b}
	\subfloat[Our method \label{distribution-comparison-b}]{%
		\includegraphics[width=0.23 \textwidth]{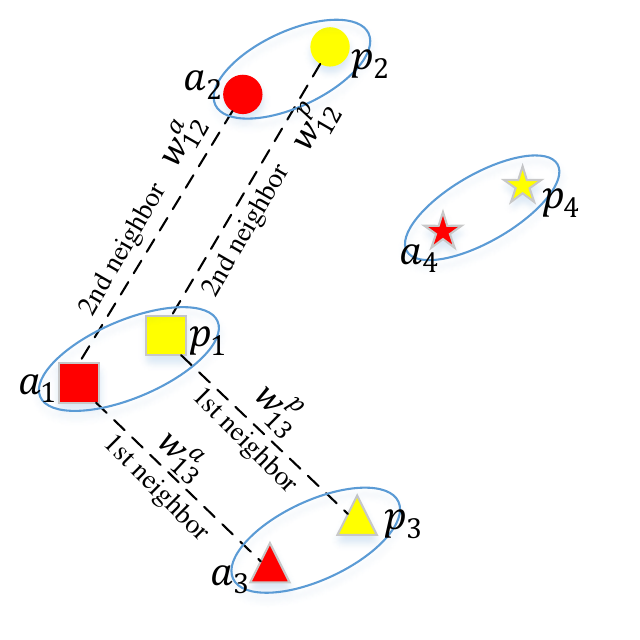}
	}
	% \vspace{-5pt}
	\caption{Distribution of descriptors learned by (a) former triplet loss and (b) our method. In (a), there exists topology difference between two matching descriptors because former triplet loss only considers Euclidean distance between descriptors and completely neglects the neighborhood information of descriptors. In (b), our method encourages similiar linear topology between matching descriptors, which means the matching descriptors have the matching $\boldsymbol {k}$NN descriptors and the similiar linear combination weights.}
	\label{distribution-comparison}
\end{figure}

% problem and motivation
However, as shown in Fig.~\ref{distribution-comparison-a}, triplet loss of former works only considers Euclidean distance between descriptors and completely neglects the neighborhood information of descriptors, which results in the topology difference between matching descripotrs in two descriptor sets.
We note that the topology of descriptor indicates the linear relationship among descriptor and its $k$NN descriptors.
Topology difference between matching descripotrs leads to the inconsistent distribution of descriptors in two sets, which is incompatible with the  one-to-one correspondence of descriptors in two sets.

% solutions
In this work, we propose a novel topology measure for triplet loss to learn topology consistent descriptors.
We first establish a topology vector for each descriptor followed by locally linear embedding (LLE)~\cite{LLE}, a common manifold learning method for dimensionality reduction, while this topology vector depicts the linear topology among the descriptor and its $k$NN descriptors.
Then we take the $l1$ distance of descriptors' topology vectors as their topology distance to indicate the neighborhood difference between descriptors.
Last we modify the the positive sample distance in the triplet loss as the dynamic weighting of Euclidean distance and topology distanfe of matching descriptors.
The consistent topology between matching descriptors is encouraged with their topology distance minimized.

% conslusions
Compared with former triplet loss, our method learns more robust descriptors since we take additional $k$NN descriptors of matching descriptors for CNN's back-propagation.
Otherwise, our method modifies and consummates the distance measure of positive samples for trieplt loss, which means our method can improve performance of many other algorithms of learning descriptors using triplet loss.
The generalization of our method is verfied in several benchmarks in Section~\ref{experiments}.

%contributions
The contributions of this paper are three-fold:
\begin{itemize}
	\item We establish a novel topology vector for each descriptor followed by LLE~\cite{LLE} and define the topology distance between descriptors to indicate their neighborhood difference.
	\item We employ the dynamic weighting strategy to fuse Euclidean distance and topology distance of matching descriptors and take the fusion result as the positive sample distance in the triplet loss.
	\item The experimental results verify the generalization of our method. We test our method on the basis of HardNet~\cite{HardNet} and CDF~\cite{DSM}, and experimental results show our method can improve their performance in several benchmarks.
\end{itemize}

\section{Related Work} \label{relatedworks}

In this section, we begin by discussing the related work in the image hashing domain, with main focus towards the motivation behind adversarial autoencoders. Then, we continue our discussion on  adversarial learning and their limitations, especially on their generalization property when matching to the target distribution (or sample complexity requirement).

\subsection{Learning-based Descriptors}
\label{LbD}
% preface
Perhaps SIFT~\cite{SIFT} is the most successful and widely used handcrafted descriptor, however, all handcrafted descriptors, including SIFT~\cite{SIFT}, LIOP~\cite{LIOP}, GLOHP~\cite{GLOH}, DAISYP~\cite{DAISY}, DSP-SIFTP~\cite{DSP-SIFTP} and BRIEF~\cite{Brief} are not robust enough as they only consider the pixel-level information instead of the high-level semantic information.
In the past several years, learning-based descriptors outperforms than handcrafted descriptors in image matching~\cite{D2_Net,RF_Net} and image retrieval~\cite{wei2017selective,gu2019towards} benefitting from powerful semantic representation of CNN.

% L2-Net
L2-Net~\cite{L2Net} proposes a CNN architecture with 7 convolutional layers and a Local Response Normalization layer to normalize descriptors, and this architecture is employed by many works~\cite{HardNet, SOSNet, DSM} including ours.
% HardNet
HardNet~\cite{HardNet} implements a hard negative mining method for learning descriptors by maximizing the nearest non-matching descriptors using triplet loss. 
% CDbin
CDbin~\cite{CDbin} combines triplet loss and other three losses for learning descriptors and explores the performance of descriptors with different lengths.
% SOSNet
SOSNet~\cite{SOSNet} proposes the Second Order Similarity Regularization in the basis of triplet loss to learn more compact descriptors.
% ELDesc
Exp-TLoss~\cite{ExpLoss} modifies triplet loss and proposes a novel exponential losses to mine harder positive samples as focal loss~\cite{focal_loss}.
% DSM
CDF~\cite{DSM} replaces the hard margin with a non-parametric soft margin with the dynamic triplet weighting to avoid the sub-optimal results.
% our method
However, all of them fail to maintain the similiar topology between matching descriptors as our work, which contributes to the more robust descriptors.

\subsection{Triplet Loss}
\label{TL}
Triplet loss consists of three parts: margin, distance of positive samples and distance of negtive samples, which updates networks by encouraging distance of negtive samples is a margin larger than distance of positive samples.

FaceNet~\cite{FaceNet} first proposes triplet loss and applies it in face recognition.
Alexander~\cite{hermans2017defense} implements the hard trpilets mining method for Person Re-Identification, which defines the hardest positive sample as positive sample with the largest distance and define the hardest negtive sample as negtive sample with smallest distance.
Wang~\cite{wang2018learning} combines the triplet loss and softmax loss to learn more discriminative features for Person Re-Identification.
Otherwise, triplet loss has been proven effective in image retrieval~\cite{lin2019tc} and learning descriptors~\cite{HardNet,ExpLoss,DSM,SOSNet}.

However, former triplet loss takes Euclidean distance between samples as the only measure, which completely neglects the toplogy of samples.
In this work, we propose a novel topology measure for triplet loss with considering neighborhood information of positive samples and verify its effectiveness on learning descriptors.

\subsection{Manifold Learning}
\label{ML}
Manifold Learning~\cite{LLE,ISOmap,LE,sLLE,hLLE,mLLE} is a commonly used dimensionality reduction method which tries to keep similiar manifold between high-dimensional data and low-dimensional data.
%In the view of manifold learning, distribution of data in high-dimensional space is complex but it satisfies the property of Euclidean space in a small local region.
% manifold learning algorithms
%LE
Laplacian Eigenmaps~\cite{LE} tries to preserve the graph structure of high-dimensional data in low-dimensional data using spectral techniques.
% ISOmap
ISOmap~\cite{ISOmap} encourages high-dimensional data and low-dimensional data have the same geodesic distance instead of Euclidean distance, where the geodesic distance means the shortest path connecting two data sample in its $k$NN graph.
% LLE
Compared with ISOmap using the global information, LLE~\cite{LLE} only tries to keep the similiar locally linear combination weight between high-dimensional data and low-dimensional data. Undoubtedly ISOmap is nuch more time-consuming.

% recent works
Manifold learning also plays an important role in recent deep learning algorithms. 
Ahmet~\cite{MoM} implements a hard training example mining method which takes manifold nearest neighbors but not Euclidean neighbors as the hard positive samples and Euclidean neighbors but not manifold nearest neighbor as the hard negtive samples.
% MMDML
Jiwen Lu~\cite{MMDML} proposes a multi-manifold deep metric learning method for image set classification by nonlinearly mapping multiple sets of image instances into a shared feature subspace.
The above methods mainly focus on image retrieval or image classification, and we are the first to introduce manifold learning into descriptors learning and image matching.

\section{Methodology}

In this section, we first review the method of learning descriptors using triplet loss in Section~\ref{Preliminaries}, and then we present the establishment of our elaborate topology vector and the difinition of topology vector in Section~\ref{TopologyMeasure}, last we illustrate the dynamic weighting strategy to fuse Euclidean distance and topology distance.

\subsection{Preliminaries}
\label{Preliminaries}
% Preliminaries
We note that learning descriptors is the image embedding from image patches to descriptor vectors. Suppose a batch of training data generates the corresponding descriptors $\chi=\{A;P\}$, where $A=\{a_1, a_2,...,a_n\},P=\{p_1, p_2,...,p_n\}$ and $n$ is the batch size. Normally descriptor vectors are unit-length and 128-dimensional as SIFT~\cite{SIFT} descriptors. Note that $a_i$ and $p_j$ are a matching pair if $i$ equals $j$ and non-matching pair otherwise.

% triplet loss
The triplet loss~\cite{FaceNet} encourages the distance of negtive samples is a margin larger than that of positive samples, which denote the non-matching pairs and matching pairs respectively in descriptors learning:
\begin{equation}
L_{triplet}= \frac{1}{n}\sum_{i=1}^n max(0,margin+ \Gamma^+(a_i,p_i) - \Gamma^-(a_i,p_i)). \label{triplet_loss}
\end{equation}
% HardNet
HardNet~\cite{HardNet} first introduces triplet loss to descriptors learning whcih tries to minimize the Euclidean distance between matching descriptors and maximize that of nearest non-matching descriptors. In HardNet,
\begin{equation}
\Gamma^+(a_i,p_i)=d_E(a_i,p_i)
\end{equation}
\begin{equation}
\Gamma^-(a_i,p_i)=min(d_E(a_i,p_{j_{min}}),d_E(a_{k_{min}},p_i))
\end{equation}
where $d_E$ is the Euclidean distance, $d_E(a_i,p_{j_{min}})$ and $d_E(a_{k_{min}},p_i)$ denote the Euclidean distance of nearest non-matching descriptors. It would be time-consuming to compute the Euclidean distance for a large number of descriptors, fortunately the dot product can be used to calculate Euclidean distance between two descriptors when descriptors are unit-length vector($\parallel a_i \parallel _2 = 1$):
\begin{equation}
d_E(a_i,p_j)=\sqrt{2-2a_i^Tp_j} \label{Euclidean_distance}
\end{equation}

% shortcoming of hardnet
However, we observe that only matching descriptors $a_i$, $p_i$ and nearest non-matching descriptor $p_{j_{min}}$ or $a_{k_{min}}$ are used for CNN's back-propagation, which leads the inconsistent distribution of descriptors in $A$ and $P$, also known as topology difference between $A$ and $P$. Actually, descriptors set $A$ and $P$ should have similiar topology because descriptors in them have a one-to-one matching relationship. In next sections we would illustrate how to reduce topology difference between $A$ and $P$.

\begin{figure}[t]
	\begin{center}
	\includegraphics[width=0.45\textwidth]{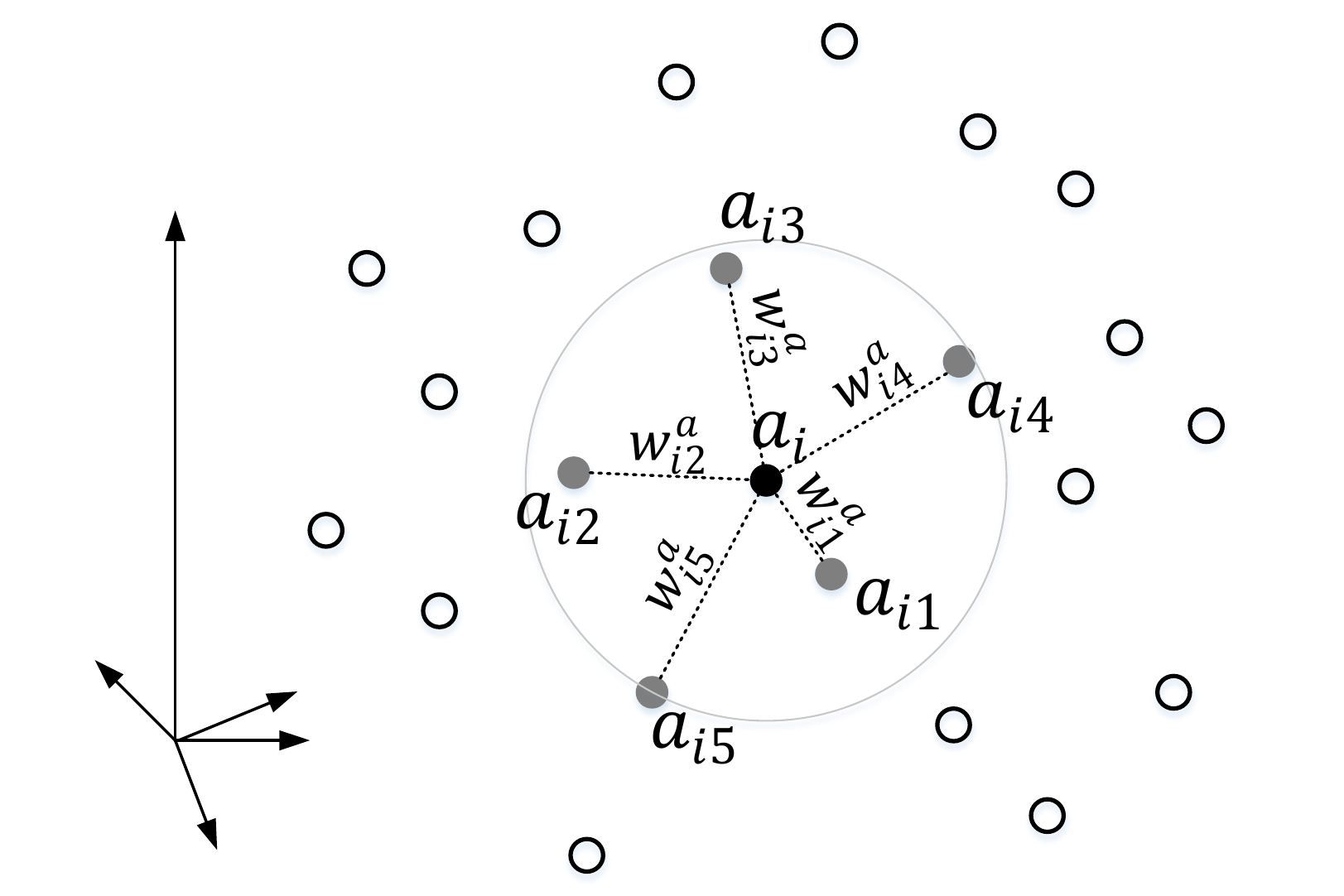}
	\end{center}
	\caption{Topological relationship among the descriptor and its kNN descriptors. In the opinion of manifold learning, property of Euclidean space is only retained in a small local region. So we first solve $\boldsymbol {k}$NN descriptors $\boldsymbol {a_{ij}}$ for $\boldsymbol {a_i}$, then we fit $\boldsymbol {a_i}$ linearly using its $\boldsymbol {k}$NN descriptors. The fitting weights $\boldsymbol {w_{ij}}$ indicate the linear topological relationship among $\boldsymbol {a_i}$ and $\boldsymbol {a_{ij}}$.}
	\label{topological_relation}
\end{figure}

\subsection{Topology Measure}
\label{TopologyMeasure}
LLE~\cite{LLE} is a common manifold learning algorithm for data dimensionality reduction, which maintains the same locally linear topology between high-dimensional data and low-dimensional data.
In the view of manifold learning, property of Euclidean space are retained in a small local region, so LLE fits each data sample by its $k$NN samples:
\begin{equation}
x_i=w_{i1}x_{i1}+w_{i2}x_{i2}+...+w_{ik}x_{ik}
\end{equation}
where $x_{ij}, \ j=1,2,...,k$ is the $k$NN samples of $x_i$, and $w_{ij}$ is the fitting weights.
Followed by LLE, we establish a locally linear topology vector for each descriptor depicting linear topological relationship among descriptor $a_i$ or $p_i$ and its kNN descriptors $a_{ij}$ or $p_{ij}$. 

% distance computing
Here we take descriptors set $A$ and its elements $a_i, \ i=1,2,...,n$ as expample, obviously we can solve this for $P$ by the same steps.
As shown in Fig.~\ref{topological_relation}, to solve linear topological relationship between $a_i$ and its kNN descriptors, we first determine kNN descriptors $a_{ij}$ for $a_i$. We compute the Euclidean distance between $a_i$ and all other descriptors in a mini-batch by Eq.~\ref{Euclidean_distance}, then we sort the distances in ascending order and take the elements corresponding to front $k$ distances as the $k$NN descriptors of $a_i$.

% fitting
The next step is to linearly fit $a_i$ using $a_{ij}$, which can be written as $a_i=w_{i1}^a a_{i1}+w_{i2}^a a_{i2}+...+w_{ik}^a a_{ik}$. So the optimization goal is:
\begin{equation}
\begin{gathered}
\mathop{\arg\min}_{w}  \| a_i-\sum_{j=1}^k w_{ij}^a a_{ij} \|^2 \\
s.t. \ \sum_{j=1}^k w_{ij}^a=1  \label{optimization}
\end{gathered}
\end{equation}
Now write the above formula in matrix form. Assume $A_i \in R^{128 \times k}$ is a matrix by $a_i$ repeating $k$ times, and $N_i^a \in R^{128 \times k}$ consists of $a_{ij}$. Now note $S_i=(A_i-N_i^a)^T(A_i-N_i^a)$, where $S_i$ is a real symmetric and semi-definite matrix, so the above optimization formula can be written as:
\begin{equation}
\begin{gathered}
\mathop{\arg\min}_{W} {W_i^a}^T S_i W_i^a \\
s.t. \ \ {W_i^a}^T \boldsymbol {1}_k=1    \label{fitting_W}
\end{gathered}
\end{equation}

This above optimization problem has the closed solution:
\begin{equation}
W_i^a=\frac{S_i^{-1} \boldsymbol {1}_k}{\boldsymbol {1}_k^T S_i^{-1} \boldsymbol {1}_k}  \label{fitting_weight}
\end{equation}
Obviously $W_i^a=[w_{i1}^a,w_{i2}^a,...,w_{ik}^a] \in R^{k}$ is the weight sequence
depicting the linear topological relationship among $a_i$ and its $k$NN descriptors $a_{ij}$. Now we expand $W_i$ to the locally linear topology vector $T_i^a=[t_{i1}^a,t_{i2}^a,...,t_{in}^a] \in R^{n}$ by the following principle:
\begin{equation}
{t^a_{ij}}=\left\{
\begin{aligned}
&w_{ij}^a, \ \ \ a_j \in kNN(a_i)\\
&0, \ \ \ \ \ \ \ otherwise\\
\end{aligned}
\right.
\end{equation}
By above equation, $t^a_{ij}$ equals $0$ if $a_j$ is not one of $k$NN descriptors of $a_i$.
Obviously we can establish the topology vector $T_i^p$ for each $p_i$ in descriptors set $P$ followed by above steps.

% conclusions
$T_i^a$ and $T_i^p$ are the topology vectors of descriptors $a_i$ and $p_i$, which depict the locally linear relationship among $a_i$ or $p_i$ and $a_{ij}$ or $p_{ij}$. The length of $T_i^a$ and $T_i^p$ is not a fixed number, while it equals batch size $n$, an important hyper-parameter of CNN. The topology vectors $T_i^a$ and $T_i^p$ are sparse arrays with $k$ non-zero elements, where $k$ is far less than $n$. Otherwise, sum of all elements in topology vectors equals $\boldsymbol 1$ by Eq.~\ref{optimization}.

We could solve a topology vector $T_i^a$ or $T_j^p$ for each descriptor $a_i$ or $p_j$ followed by above steps, then we take the $l1$ distance between $T_i^a$ and $T_j^p$ as the topology distance between descriptors $a_i$ and $p_j$:
\begin{equation}
d_T(a_i, p_j)=\frac{1}{4} \| T_i^a - T_j^p \|_1 \label{topology_distance}
\end{equation}
Note that $4$ is the maximun value of $\| T_i^a - T_j^p \|_1$, which normalizes the topology distance into the range of 0 to 1. Meanwhile, we choose the $l_1$ distance to measure difference between topology vectors as they are sparse vectors. 

The topology distance $d_T(a_i, p_i)$ reflects the neighborhood difference between matching descriptors $a_i$ and $p_i$, while two aspects are required by a small topology distance: $k$NN descriptors of $a_i$ match that of $p_i$ and fitting weights $w_{ij}^a$ are similiar with $w_{ij}^p$.
We hope the matching descriptors $a_i$ and $p_i$ have the consistent local topology so that global topology difference between $A$ and $P$ is small.

\subsection{Dynamic Weighting Strategy}
\label{DynamicWeightingStrategy}
In HardNet~\cite{HardNet}, only the matching descriptors and nearest non-matching descriptors are used for for CNN's back-propogation, which neglects the topology similarity between descriptor sets $A$ and $P$.
In this section we encourage the similiar topology between matching descriptors $a_i$ and $p_i$ by minimizing their topology distance $d_T(a_i, p_i)$.
So we define the distance of positive samples $\Gamma^+(a_i,p_i)$ in triplet loss as following:
\begin{equation}
\Gamma^+(a_i,p_i)=\lambda d_E(a_i,p_i) + (1-\lambda)d_T(a_i,p_i)
\label{distance_positive}
\end{equation}
where weight $\lambda$ is a hyper-parameter in range 0 to 1 to balance the Euclidance distance and topology distance.

By minimizing $\Gamma^+(a_i,p_i)$, first we can reduce Euclidean distance of matching descriptors, and then we encourage $a_i$ and $p_i$ have the matching $k$NN descriptors, last we reduce the difference of topological weights between $a_i$ and $p_i$, while early works~\cite{HardNet,SOSNet,ExpLoss,DSM} only consider the first item.
Compared with Hardnet only using matching descriptors and nearest non-matching descriptors to update CNN, our method considers additional $k$NN descriptors of matching descriptors for CNN's back-propagation.

In Eq.~\ref{distance_positive}, weight $\lambda$ is an important parameter that directly affects the performance of descripotrs.
We note that the larger $\lambda$ focuses more on the Euclidean distance between descriptors and contributes to the more discriminative descriptors, and a smaller $\lambda$ focuses more on the topology distance between descriptors and contributes to the more robust descriptors.
In this paper, we employ the dynamic weighting strategy to to fuse the Euclidean distance and topology distance of matching descriptors. Specifically, we choose a larger $\lambda$ in the former training epochs, then we decay the value of $\lambda$ gradually. The value of $\lambda$ in $n$-th interation can be solved by the following equation:
\begin{equation}
\lambda = {\rm max}(1- \lceil \frac{{\rm max}(0, n-n_0)}{N} \rceil \times r, 0.5)
\label{lambda}
\end{equation}
By Eq.~\ref{lambda}, $\lambda$ euqals $1$ in the initial $n_0$ iterations during training, and decays $r$ for each $N$ iterations.
The minimum value of $\lambda$ is $0.5$, which takes Euclidean distance and topology distance equally.

For the negtive samples, non-matching descriptors in triplet loss, we found there is no need to encourage the large toplogy distance for them because there may exist matching pairs inside $k$NN descriptors of non-matching descriptors. So we define the distance of negtive samples as the Euclidean distance of nearest non-matching descriptors like HardNet.

We note that our method have two overwhelming advantages compared with former triplet loss:
First, besides the point-to-point distance constraints, our method takes advantage of the high-order topology constraints to improve the robustness of descriptors;
Second, our method considers the neighborhood information of positive sample, which menas more descriptors are used to update CNN.

\section{Experiments}
\label{experiments}
The main contribution of our work is to propose the topology measure besides Euclidean distance for triplet loss to encourage the similiar topology between descriptor sets $A$ and $P$. To verify the generalization of our method, we test our method on the basis of HardNet~\cite{HardNet} and CDF~\cite{DSM}, where HardNet first introduces triplet loss into learning descriptors and CDF is the state-of-the-art method of learning descriptors using triplet loss.

To validate the performance of our topology consistence descriptors \textbf{TCDesc}, we conduct our experiments in three benchmarks: UBC PhotoTourism~\cite{UBC}, HPatches~\cite{HPatches} and W1BS dataset~\cite{WxBS}. 
UBC PhotoTourism~\cite{UBC} is currently the largest and the most widely used local image patches matching dataset, which consists of three subsets(\textit{Liberty}, \textit{Notredame} and \textit{Yosemite}) with more than 400k image patches.
HPatches~\cite{HPatches} presents the more complicated and more comprehensive three tasks to evaluate descriptors: \textit{Patch Verification}, \textit{Image Matching}, and \textit{Patch Retrieval}.
W1BS dataset~\cite{WxBS} consists of 40 image pairs and provides more challenging tasks with several nuisance factors to explore the performance of descriptors in extreme conditions.

\begin{table*}[ht]
	\centering
	\caption{Patch verification performance on the UBC PhotoTourism benchmark. Numbers shown are FPR95(\%), while the lower FPR95 indicates the better performance of learned descriptors. Plus "+" denotes training with data augmentation. We test our method on the basis of HardNet~\cite{HardNet} and CDF~\cite{DSM}, which is noted as TCDesc-HN and TCDesc-CDF respectively.}
	\begin{tabular}{cccccccccc}
		\hline
		\multirow{2}{*}{Descriptors} & \multirow{2}{*}{Length} & Train & Notredame     & Yosemite    & Liberty       & Yosemite      & Liberty      & Notredame     & \multirow{2}{*}{Mean} \\ \cline{4-9}
		&                         & Test  & \multicolumn{2}{c}{Liberty} & \multicolumn{2}{c}{Notredame} & \multicolumn{2}{c}{Yosemite} &                       \\ \hline
		SIFT~\cite{SIFT}                         & 128                     &       & \multicolumn{2}{c}{29.84}   & \multicolumn{2}{c}{22.53}     & \multicolumn{2}{c}{27.29}    & 26.55                 \\
		DeepDesc~\cite{DeepDesc}                     & 128                     &       & \multicolumn{2}{c}{10.9}    & \multicolumn{2}{c}{4.40}      & \multicolumn{2}{c}{5.69}     & 7.0                   \\
		L2-Net+~\cite{L2Net}                      & 128                     &       & 2.36          & 4.70        & 0.72          & 1.29          & 2.51         & 1.71          & 2.23                  \\
		CS L2-Net+~\cite{L2Net}                   & 256                     &       & 2.55          & 4.24        & 0.87          & 1.39          & 3.81         & 2.84          & 2.61                  \\
		HardNet~\cite{HardNet}                     & 128                     &       & 1.47          & 2.67        & 0.62          & 0.88          & 2.14         & 1.65          & 1.57                  \\
		HardNet+~\cite{HardNet}                     & 128                     &       & 1.49          & 2.51        & 0.53          & 0.78          & 1.96         & 1.84          & 1.51                  \\
		DOAP+~\cite{DOAP}                        & 128                     &       & 1.54          & 2.62        & 0.43          & 0.87          & 2.00         & 1.21          & 1.45                  \\
		DOAP-ST+~\cite{DOAP,STN}                     & 128                     &       & 1.47          & 2.29        & 0.39          & 0.78          & 1.98         & 1.35          & 1.38                  \\
		ESE~\cite{ESE}                          & 128                     &       & 1.14          & 2.16        & 0.42          & 0.73          & 2.18         & 1.51          & 1.36                  \\
		SOSNet~\cite{SOSNet}                       & 128                     &       & 1.25          & 2.84        & 0.58          & 0.87          & 1.95         & 1.25          & 1.46                  \\
		Exp-TLoss~\cite{ExpLoss}                    & 128                     &       & \textbf{1.16}          & 2.01        & 0.47          & 0.67          & 1.32         & 1.10          & 1.12                  \\
		CDF+~\cite{DSM}                          & 128                     &       & 1.21          & 2.01        & 0.39          & 0.68          & 1.51         & 1.29          & 1.18                  \\ 
		TCDesc-HN+                          & 128                     &       & 1.47          & 2.38        & 0.43          & 0.72          & 1.47         & 1.23          & 1.28                   \\
		TCDesc-CDF+                          & 128                     &       & 1.18           & \textbf{1.99}        & \textbf{0.34}          & \textbf{0.65}          & \textbf{1.26}         & \textbf{1.08}          & \textbf{1.08}                  \\ \hline
	\end{tabular}  \label{UBCbenchmark}
\end{table*}

\subsection{Implementations}
We use the same configuration as former works to guarantee the improvement of experimental results attributes to our novel topology measure.
We use the CNN architecture proposed in L2-Net~\cite{L2Net} with seven convolutional layers and a Local Response Normalization layer.
We note that we only train our network on benchmark UBC PhotoTourism and then test other two benchmarks using the trained model.
The size of image patches in UBC PhotoTourism is $64 \times 64$, then we downsample each patch to size of $32 \times 32$, which is required by of L2-Net.
We conduct data augmentation as CDF~\cite{DSM} to flip or rotate image patches randomly.
To accord with HardNet~\cite{HardNet} and CDF~\cite{DSM}, we set the training batch size to be 1024.
We train our network for $250k$ iterations using Stochastic Gradient Descent(SGD) with momentum $0.9$ and weight decay $10^{-4}$, and the learning rate is decayed linearly from $0.1$ to $0$.

There are two important hyper-parameters in our method: the number of nearest neighbor descriptors $k$ and the weight to balance Euclidean distance and toppology distance $\lambda$.
Our novel topology measure consider $k$ nearest neighbor descriptors of matching descriptor for CNN's back-propagation, so the larger $k$ means we use more descriptors to update CNN's parameters in each iteration.
However, by the opinion of maniflod learning, the property of Euclidean space is only retained in a small local region. So it's not feasible for us to define a very large $k$.
%Actually, the value of $k$ in the range of $10$ to $15$ is recommended by previous researches~\cite{hLLE,mLLE,sLLE}.
Former works~\cite{hLLE,mLLE,sLLE} choose the value of $k$ in the range of $10$ to $15$.
We set $k$ in our experiments to be $20$ considering the large batch size 1024.

In Section~\ref{DynamicWeightingStrategy}, we define the distance of positive samples in triplet as the dynamic weighting of Euclidean distance and topology distance of matching descriptors.
As shown in Eq.~\ref{lambda}, the weight $\lambda$ is determined by initial steps $n_0$, decay steps $N$ and decay rate $r$.
In our experiments, we set $n_0$, $N$ and $r$ as $5\times10^4$, $10^4$ and $0.025$ respectively.
Within the total $250k$ iterations, the weight $\lambda$ equals $1.0$ in the initial $50k$ iterations and declines $0.025$ for each $10k$ iterations in the later $200k$ iterations, which means $\lambda$ declines from $1.0$ to $0.5$ during the whole training.

\subsection{UBC PhotoTourism benchmark}
UBC PhotoTourism~\cite{UBC} is the first large benchmark of learning descriptors from image patches which consists of more than 400k image patches extracted from large 3D reconstruction scenes.
UBC PhotoTourism consists of three subsets: \textit{Liberty}, \textit{Notredame} and \textit{Yosemite}.
Usually we train one sbuset and test other two subsets.
The false positive rate at 95\% recall (FPR95) is employed by UBC PhotoTourism to evaluate the performance of learned descriptors, where the lower FPR95 indicates the better performance.

We test our method on the basis of HardNet~\cite{HardNet} and CDF~\cite{DSM}, which are the first work introducing triplet loss into learning descriptors and the state-of-the-art method of learning descriptors using triplet loss respectively.
Specifically, we modify the distance of positive sample in their triplet losses as the linear weighting of Euclidean distance and topology distance of matching descriptors.
Then we compare our method with SIFT~\cite{SIFT}, DeepDesc~\cite{DeepDesc}, L2-Net~\cite{L2Net}, HardNet~\cite{HardNet}, DOAP~\cite{DOAP}, ESE~\cite{ESE}, SOSNet~\cite{SOSNet}, Exp-TLoss~\cite{ExpLoss} and CDF~\cite{DSM}.
We present the performance of descriptors learned by various algorithms in Table.~\ref{UBCbenchmark}.

As can be seen, our novel topology measure improves performance of both descriptors learned by HardNet and CDF.
Specifically, mean FPR95 of HardNet declines from 1.51 to 1.28 after intruducing our topology measure and that of CDF declines from 1.18 to 1.08.
Furthermore, our method reduces the FPR95 of HardNet and CDF on every test task.
Otherwise, as presented in Table.~\ref{UBCbenchmark}, our TCDesc on the basis of CDF leads the state-of-the-art result with the lowest FPR95 1.08.

The experimental results on UBC PhotoTourism benchmark validate the generalization of our method: we can improve performances of several descriptors learned by former triplet loss.

\begin{figure*}[ht]

	\centering
	\renewcommand{\thesubfigure}
	\subfloat{%
		\includegraphics[width=0.32 \textwidth]{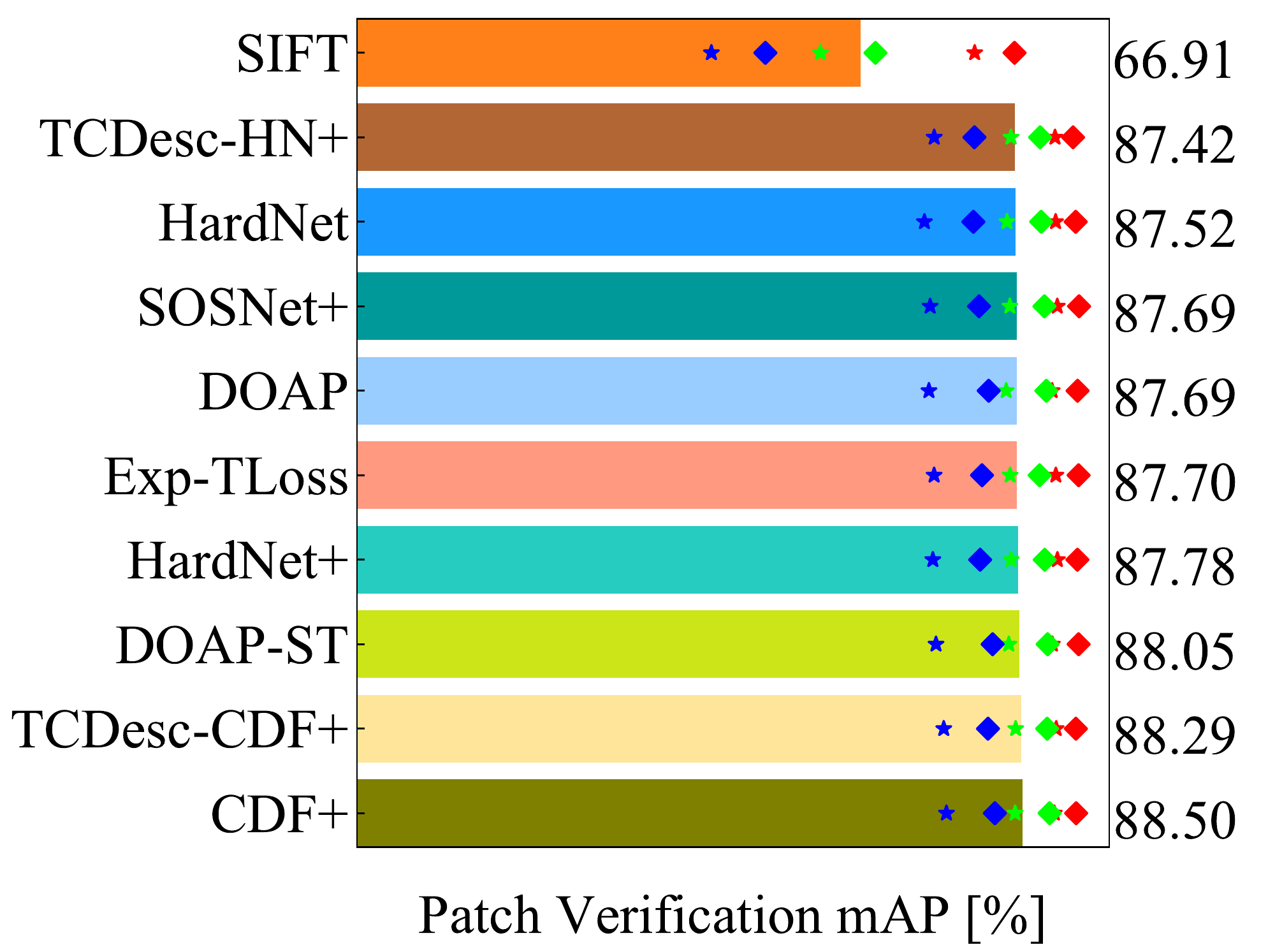}
	}
	\hfill
	\renewcommand{\thesubfigure}
	\subfloat{%
		\includegraphics[width=0.32 \textwidth]{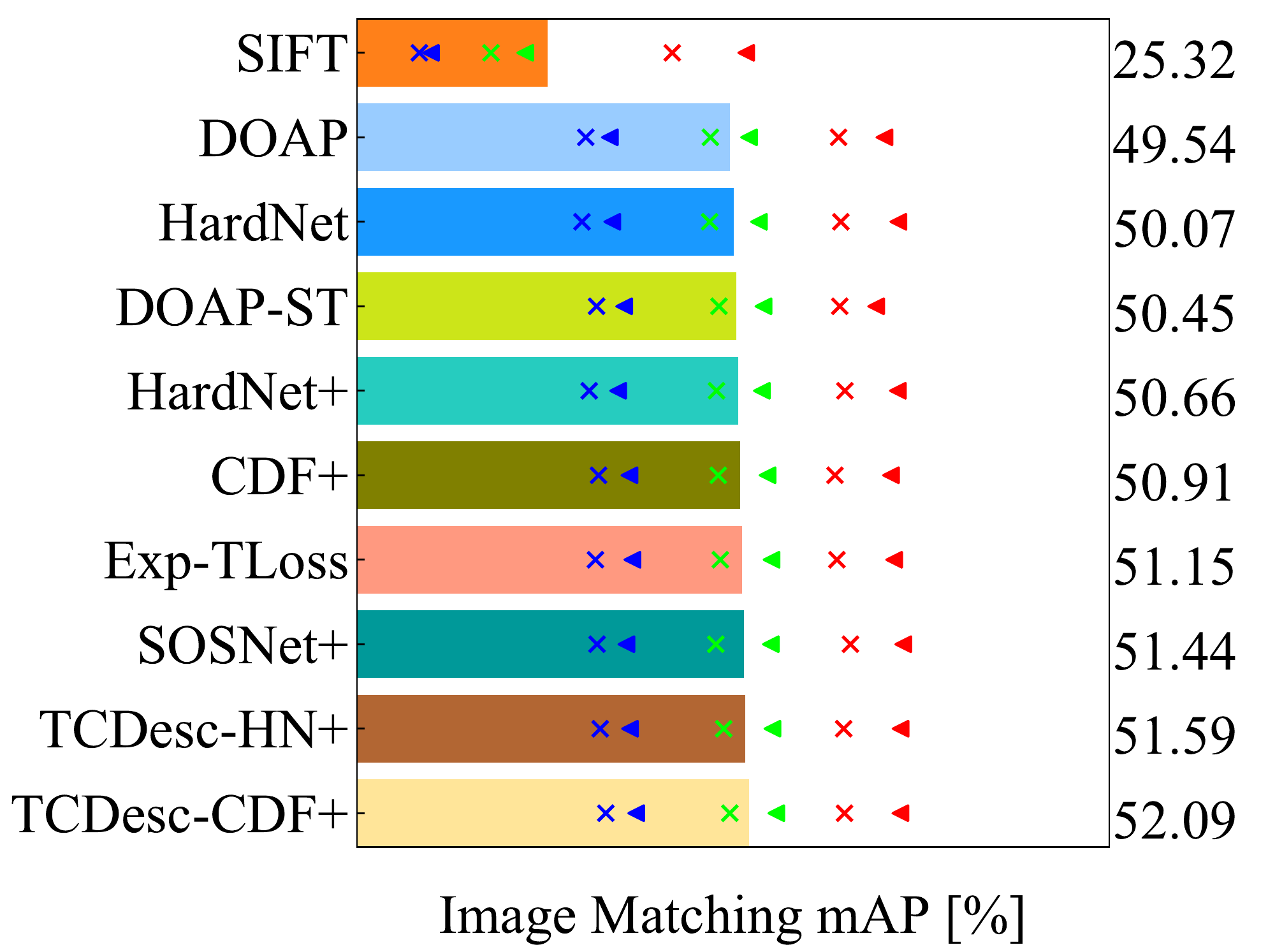}
	}
	\hfill
		\renewcommand{\thesubfigure}
	\subfloat{%
		\includegraphics[width=0.32 \textwidth]{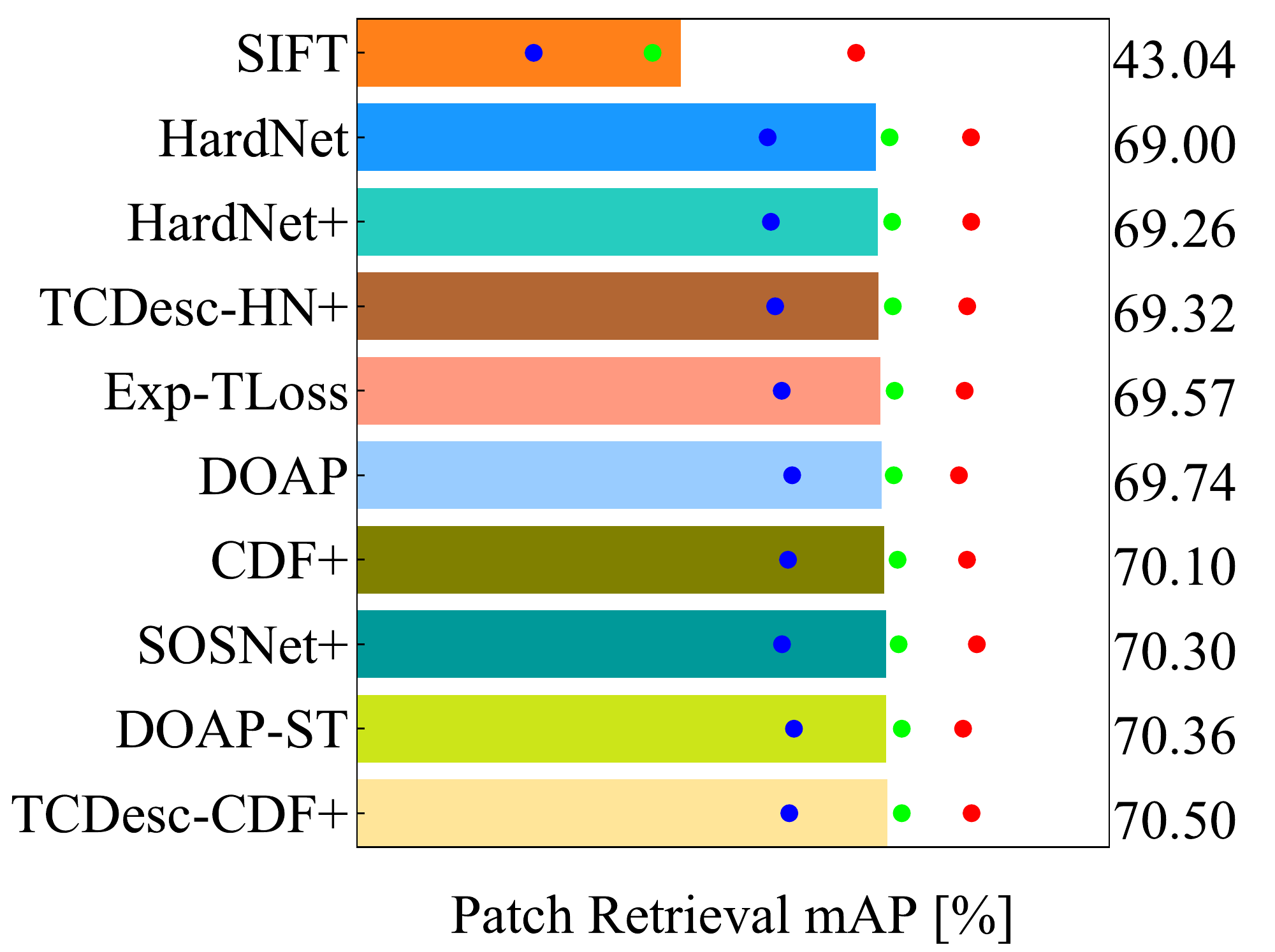}
	}
	\hfill
	% \vspace{-5pt}
	\caption{Performance of descriptors on HPatches benchmark. In these three figures, colors of markers indicate the difficulty level of tasks: easy (red), hard (green), tough (blue). In the left figure, DIFFSEQ($\Diamond$) and SAMESEQ($\star$) represent the source of negative examples in verification task. In the middle figure,  ILLUM ($\times$) and VIEWPT ($\triangleleft$) indicate the influence of illumination and viewpoint changes in matching task. All the descriptors are generated by the model trained on subsets \textit{Liberty} of UBC PhotoTourism benchmark.}
	\label{HPatches}
\end{figure*}

\subsection{HPatches benchmark}
HPatches benchmark~\cite{HPatches} consists of 116 sequences where the main nuisance factor of 57 sequences is illumination and that of 59 sequences is viewpoint.
Feature points in the 3D scenes are detected by DoG, Hessian-Hessian and Harris-Laplace. Then the reference feature points are projected to the target image using the groundtruth homographies to solve the target feature points.

Compared with UBC PhotoTourism benchmark, HPatches benchmark~\cite{HPatches} provides more diverse data samples and more sophisticated tasks.
HPatches~\cite{HPatches} defines three tasks to evaluate descriptors: \textit{Patch Verification}, \textit{Image Matching}, and \textit{Patch Retrieval}, and  each task is categorized as "Easy", "Hard" or "Tough" according to the amount of geometric noise or changes in viewpoint and light illumination.
The mean average precision(mAP) is employed to evaluate descriptors and the higher mAP indicates the better performance.

We use model trained on subsets \textit{Liberty} of UBC PhotoTourism benchmark to generate descriptors from image patches of HPatches.
We compare our topology consistent descriptors \textbf{TCDesc-HN} and \textbf{TCDesc-CDF} with SIFT~\cite{SIFT}, HardNet~\cite{HardNet}, DOAP~\cite{DOAP}, SOSNet~\cite{SOSNet}, Exp-TLoss~\cite{ExpLoss} and CDF~\cite{DSM}, where our descriptors TCDesc-HN and TCDesc-CDF are trained on the basis of HardNet~\cite{HardNet} and CDF~\cite{DSM} respectively.

As can be seen in Fig.~\ref{HPatches}, there only exists a small margin among mAP of various learning-based descriptors in three tasks.
In task \textit{Patch Verification}, our TCDesc-CDF performs a little worse than CDF, and TCDesc-HN performs a little worse than HardNet, which mainly results from the topology difference of descriptors in benchmarks UBC PhotoTourism and HPatches.
In task \textit{Image Matching}, our TCDesc-CDF and TCDesc-HN lead the state-of-the-art results and perform much better than CDF and TCDesc-HN, which proves the effectiveness of our topology consistent descriptors in image matching.
In task \textit{Patch Retrieval}, our TCDesc-CDF and TCDesc-HN both outperform than CDF and TCDesc-HN, and the TCDesc-CDF achieves the highest mAP(70.50) in this task.

\begin{figure*}[ht]
	\begin{center}
		\includegraphics[width=1.0\textwidth]{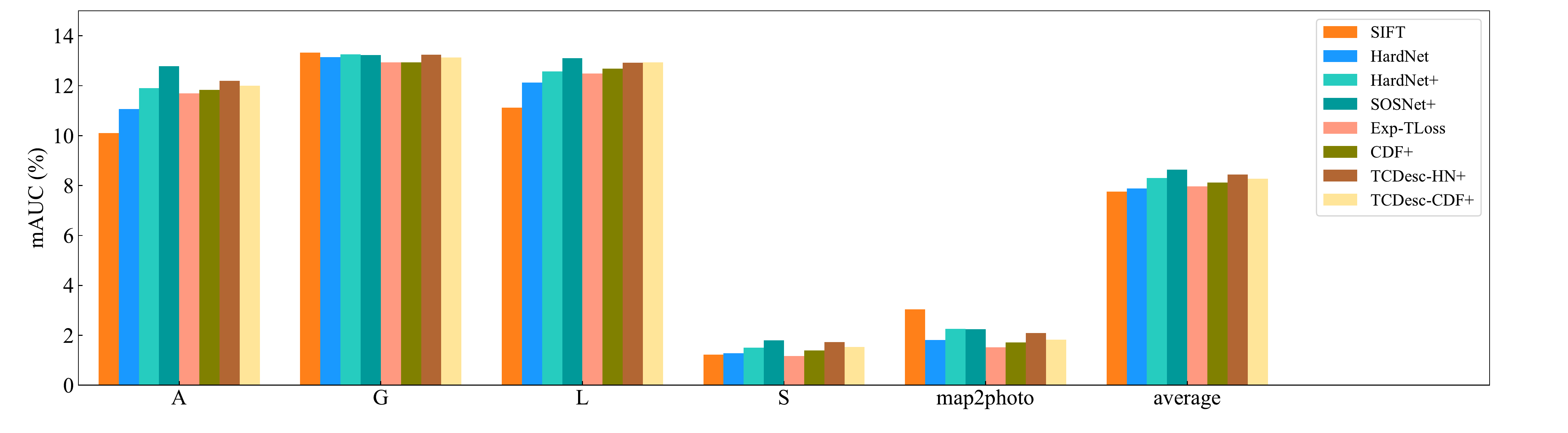}
	\end{center}
	\caption{Descriptor evaluation on the W1BS patch dataset. W1BS dataset consists of 40 image pairs divided into 5 parts by the nuisance factor: Appearance(A), Geometry(G), Illumination(L), Sensor(S) and Map to photo. The larger mAUC indicates the better performance of descriptors.}
	\label{wxbs}
\end{figure*}

\subsection{Wide baseline stereo}
Wide baseline stereo matching~\cite{wide_baseline} aims to find correspondences of two images in wide baseline setups, i.e., cameras with distant focal centers. So it is more challenging than normal image matching.
To verify generalization of our \textbf{TCDesc} and prove its advantages in extreme conditions, we conduct our experiments on W1BS benchmark~\cite{WxBS}.

W1BS dataset consists of 40 image pairs divided into 5 parts by the nuisance factor:
\\
\textit{Appearance(A):} difference in object appearance caused by season or weather changes;
\\
\textit{Geometry(G):} difference in camera positions and scales;
\\
\textit{Illumination(L):} difference in direction, intensity and wavelength of light sources;
\\
\textit{Sensor(S):} difference in sensor type, including visible, IR, MR;
\\
\textit{Map to photo:} object image and map image.

W1BS datase uses multi detectors MSER~\cite{MSER}, Hessian-Affine~\cite{Hessian-Affine} and FOCI~\cite{FOCI} to detect affine-covariant regions and normalize the regions to size $41 \times 41$.
The average recall on ground truth correspondences of image pairs are employed to evaluate the performance of descriptors.

We compare our \textbf{TCDesc-HN} and \textbf{TCDesc-CDF} with SIFT~\cite{SIFT}, HardNet~\cite{HardNet}, SOSNet~\cite{SOSNet}, Exp-TLoss~\cite{ExpLoss} and CDF~\cite{DSM}.
Like the former experiment, we use the model trained on subsets \textit{Liberty} of UBC PhotoTourism benchmark to generate descriptors.
The experimental results are presented in Fig.~\ref{wxbs} where the larger mAUC indicates the better performance.
The average mAUC of our TCDesc-HN is 8.44\%, which is larger than that of HardNet 8.30\%; The average mAUC of our TCDesc-CDF is 8.28\%, and it is larger than that of CDF 8.12\%.
Conclusion could be drawn that our method can also improve performance of descriptors learned by triplet loss in extreme condition.

\begin{table}[t]
	\centering
	\caption{Impact of hyper-parameter $\boldsymbol {k}$. The larger $\boldsymbol {k}$ means that we define a larger local region to depict linear topology for descriptors and take more descriptors for CNN's back-propagation. However, we found that descriptors perform similiarly under different values of $\boldsymbol {k}$.}
	\begin{tabular}{cccccc}
		\hline
		\multirow{2}{*}{parameter} & \multirow{2}{*}{value} & train & \multicolumn{2}{c}{Liberty} & \multirow{2}{*}{Mean} \\ \cline{4-5}
		&                        & test  & Notredime     & Yosemite    &                       \\ \hline
		\multirow{4}{*}{k}    & 5                      &       & 0.38          & 1.27        & 0.83                  \\
		& 10                     &       & 0.37          & \textbf{1.21}        & \textbf{0.79}                  \\
		& 15                     &       & 0.39          & 1.30        & 0.85                  \\
		& 20                     &       & \textbf{0.34}          & 1.26        & 0.80                  \\ \hline
	\end{tabular}  \label{UBC_k}
\end{table}

\begin{figure*}[t]
	
	\centering
	\renewcommand{\thesubfigure}
	\subfloat{%
		\includegraphics[width=0.32 \textwidth]{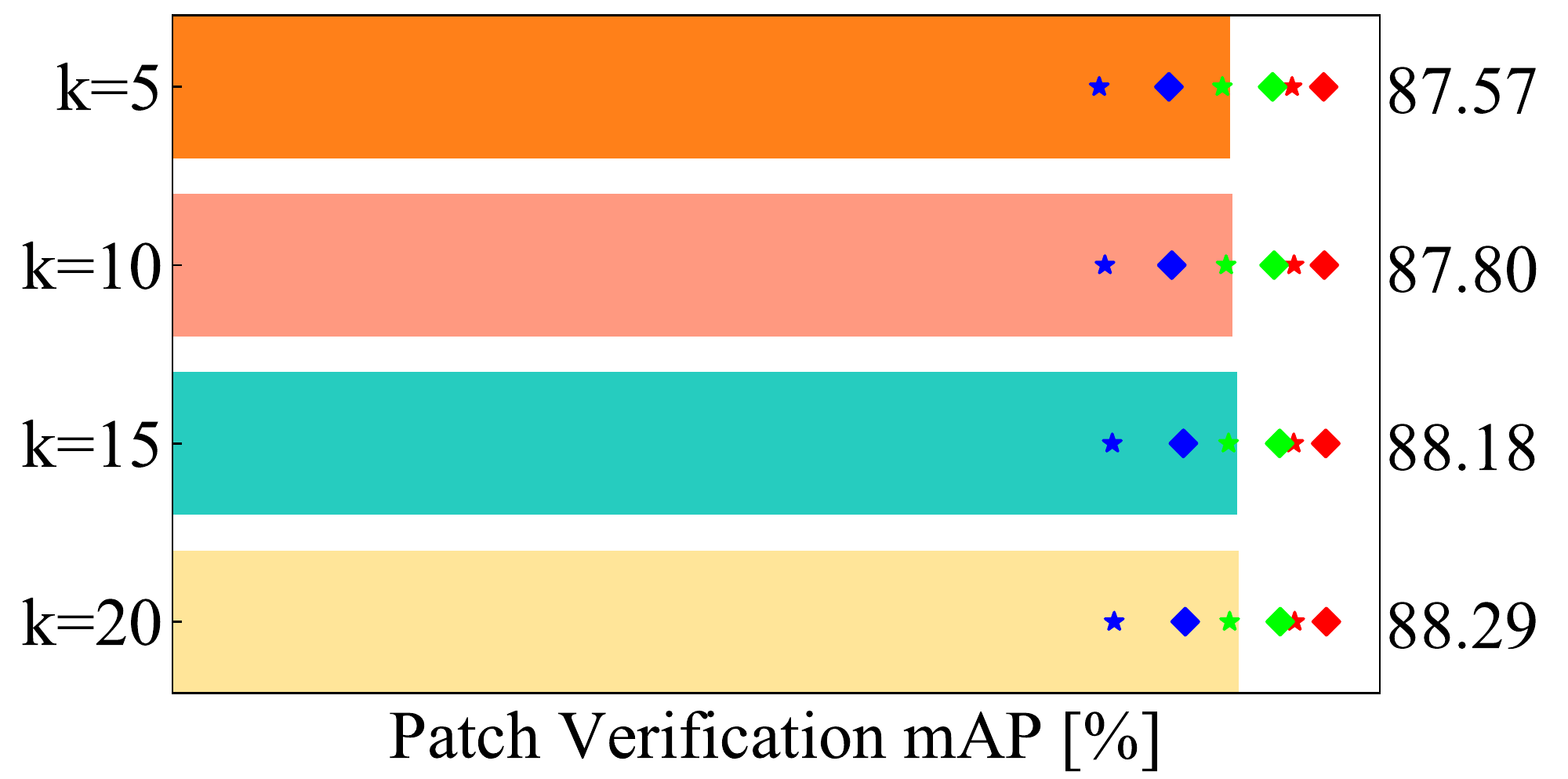}
	}
	\hfill
	\renewcommand{\thesubfigure}
	\subfloat{%
		\includegraphics[width=0.32 \textwidth]{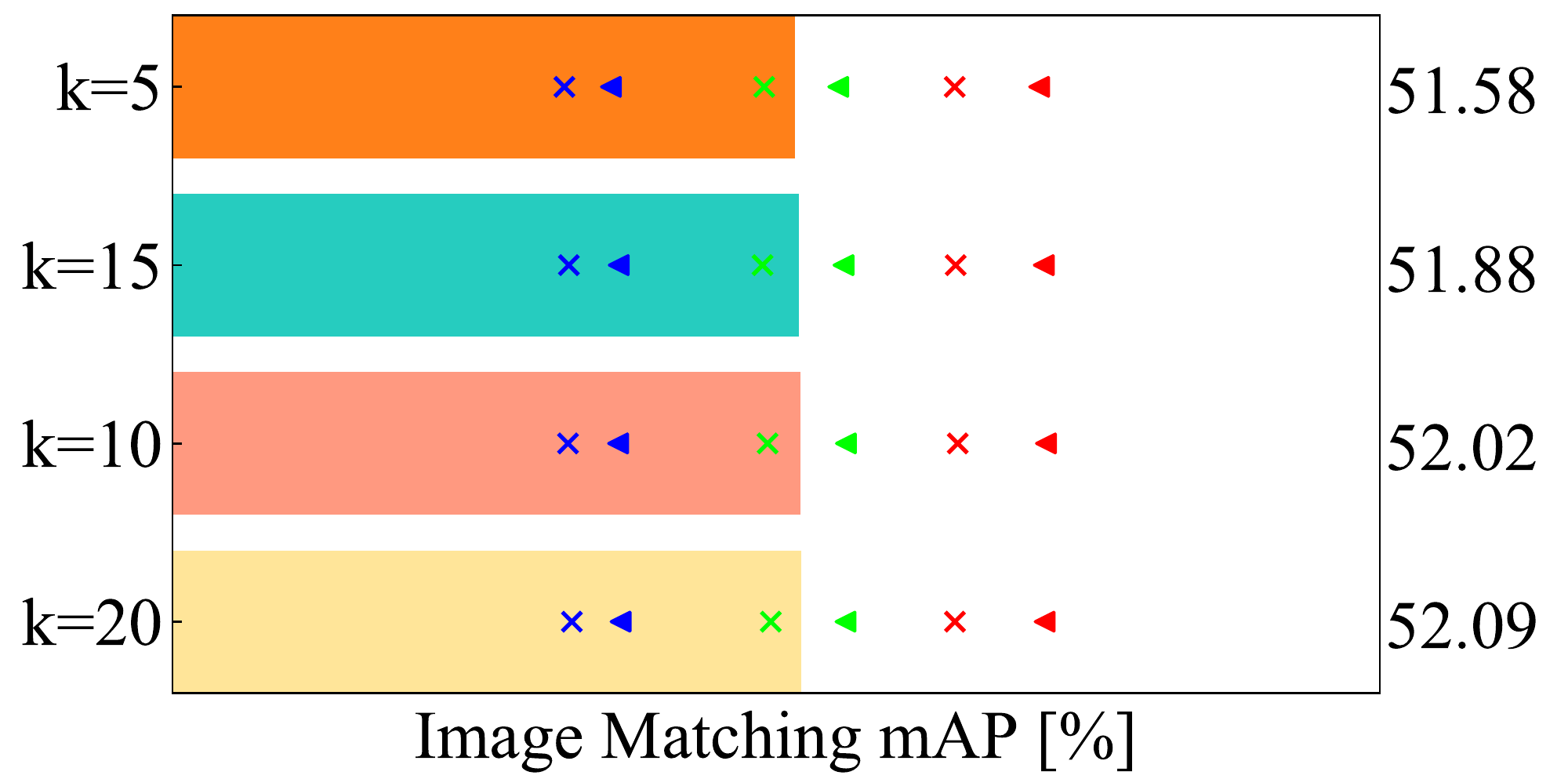}
	}
	\hfill
	\renewcommand{\thesubfigure}
	\subfloat{%
		\includegraphics[width=0.32 \textwidth]{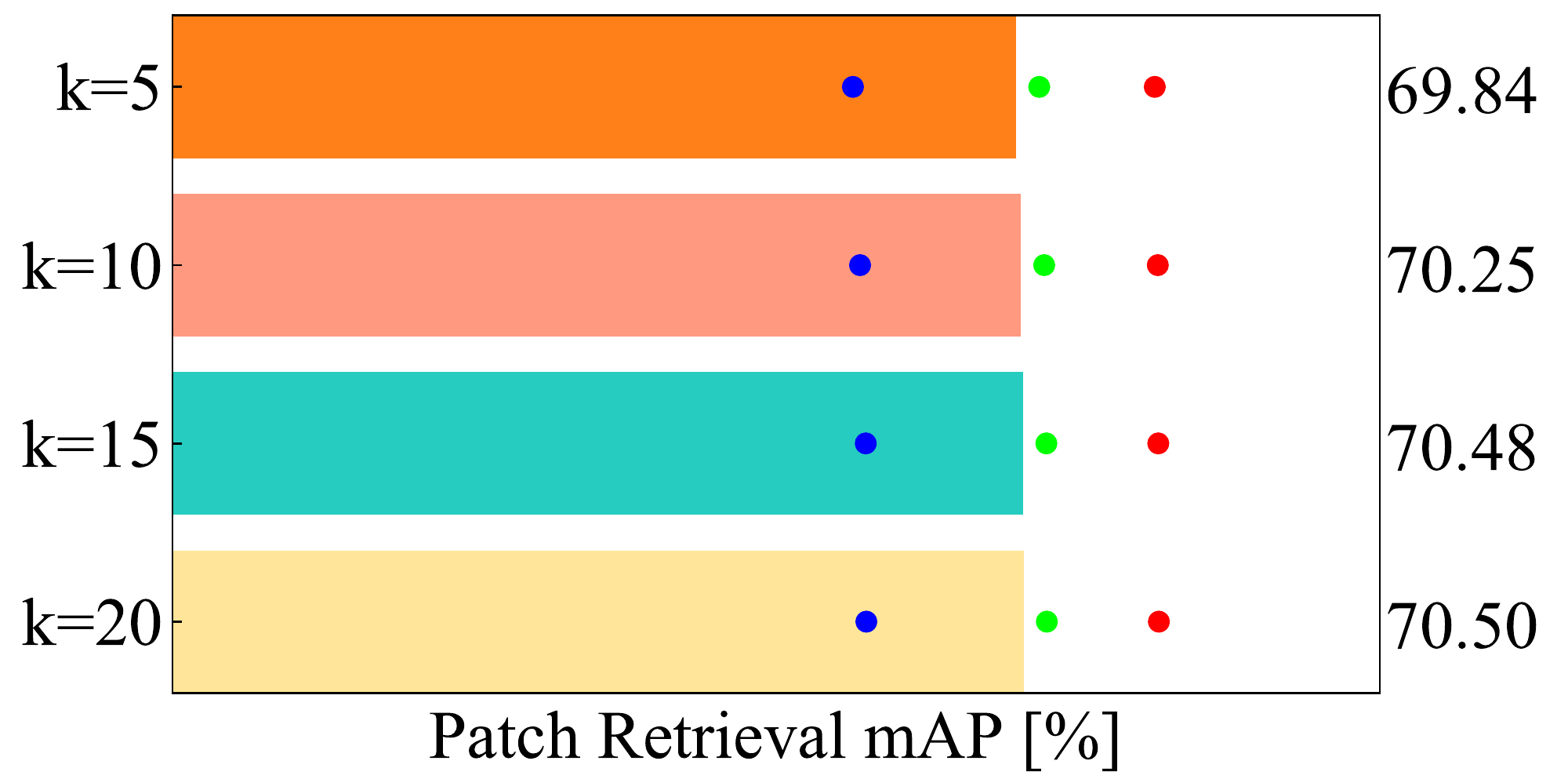}
	}
	\hfill
	% \vspace{-5pt}
	\caption{Performance of TCDesc-CDF on benchmark HPatches under different $\boldsymbol {k}$. We use the model trained on \textit{Liberty} of UBC PhotoTourism to generate the descriptors. As can be seen, the larger $\boldsymbol {k}$ contributes the better performance though their performances on UBC PhotoTourism are similiar.}
	\label{HPatches_k}
\end{figure*}

\section{Discussions}
In this Section, we explore the impact of hyper-parameters $k$ to our topology consistent descriptors \textbf{TCDesc-CDF}. 
We first train our models on subset \textit{Liberty} of UBC PhotoTourism benchmark and test in other two subsets under different values of $k$.
The larger $k$ means that we define a larger local region to depict linear topology for descriptors and take more descriptors for CNN's back-propagation.
However, we found that descriptors perform similiarly under different values of $k$ on UBC PhotoTourism.

We then conduct our experiment on HPatches benchmark.
We evaluate the performances of descriptors generated by models in Table~\ref{UBC_k}.
As can be seen in Fig~\ref{HPatches_k}, the larger $k$ contributes the better performance in task \textit{Patch Verification} and \textit{Patch Retrieval}.
In task \textit{Image Matching}, descriptors under $k$ of 10 outperform than descriptors under $k$ of 15, which may result from the worse model with larger FPR95 as presented in Table~\ref{UBC_k}.
We conclude that the larger $k$ contributes to the more robust descriptors: descriptors generated by the trained model under large $k$ performs better than that under smalle $k$ on HPatches benchmark, though they perform similiarly on UBC PhotoTourism benchmark.

\section{Conclusions}
We observe the former triplet loss fails to maintain the similiar topology between two descriptor sets since it takes the Euclidean distance between descriptors as the only measure.
In this work, we propose a novel topology measure to learn topology consistent descriptors.
Inspired by LLE, we first construct a topology vector for each descriptor which decipts the linear topology relationship among descriptor and its $k$NN descriptors.
Then we define the topology distance of descriptors as the difference of their topology vector, where the topology distance indicates the neighborhood difference of descriptors.
Last we employ the dynamic weighting strategy to fuse the Euclidean distance and topology distance of matching descriptors modify the distance of positive samples of triplet loss as the fusion result.
The similiar topology between two descriptor sets are encouraged with topology distance of matching descriptors minimized.

Experimental results on several benchmarks validate the generalization of our method since our method can improve performance of several algorithms using triplet loss. Last we discuss the impact of hyper-parameter $k$ and found the larger $k$ contributes the more robust descriptors.

However, our method is not appropriate for learning binary descriptors because the binary descriptor can not be linear fitted by its $k$NN descriptors with float fitting weights.
We note that the idea of our method, locally linear topology consistency can be extended to many other fields of image embedding, such as face recognition, person ReID, image retrieval.

\newpage


% \begin{acks}
% Will be added upon the acceptance of this paper.

% \end{acks}

%%
\balance
% \newpage
\bibliographystyle{ACM-Reference-Format}
\bibliography{descriptors}


\end{document}